\documentclass[10pt,journal]{IEEEtran} 
\usepackage{epsfig}
\usepackage{subfigure}
\usepackage{amsmath,amsfonts}
\usepackage{color}
\usepackage{pstricks}
\usepackage{graphicx}
\usepackage{rotating}

\usepackage{psboxit}

\begin{document}

\setcounter{page}{1}

\title{Delta divergence: A novel decision cognizant measure of classifier incongruence}

\author{
Josef Kittler~\IEEEmembership{Life Member,~IEEE} and Cemre Zor

\thanks{The authors are with the
Centre for Vision, Speech and Signal Processing, University of
Surrey, Guildford, GU2 7XH, UK. E-mail: \{J.Kittler, C.Zor\}@surrey.ac.uk 

This work was carried out as part of EPSRC project "Signal processing in a networked battlespace"  under contract EP/K014307/1
. The EPSRC financial support is gratefully acknowledged. } }

%
\maketitle
\thispagestyle{plain}\pagestyle{plain}
\begin{abstract}
Disagreement between two classifiers regarding the class membership of an observation in pattern recognition can be indicative of an anomaly and its nuance. As in general classifiers base their decision on class aposteriori probabilities, the most natural approach to detecting classifier incongruence is to use divergence. However, existing divergences are not particularly suitable to gauge classifier incongruence. In this paper, we postulate the properties that a divergence measure should satisfy and propose a novel divergence measure, referred to as {\it Delta divergence}. In contrast to existing measures, it is decision cognizant. The focus in Delta divergence on the dominant hypotheses has a clutter reducing property, the significance of which grows with increasing number of classes. The proposed measure satisfies other important properties such as symmetry, and independence of classifier confidence. The relationship of the proposed divergence to some baseline measures is demonstrated experimentally, showing its superiority.   

\end{abstract}


\section{Introduction}
\label{intro_info}

Divergence in information theory has been intensively studied and researched over the last six decades. On one hand the massive interest in the subject has been driven by the diversity of applications where divergence plays the key role  as an objective function. On the other hand the investigation  of the underlying theoretical properties of divergence  has motivated the discovery of new measures with tailor made characteristics that are fine tuned for specific applications. This dual drive has produced extensive families of divergences which are encapsulated in the generic expressions presented e.g.  in \cite{Girardin-2015}, with many specific examples listed in the review paper, e.g.  \cite{Liese-2006}. We shall provide a very brief overview of these developments in Section \ref{survey} and give representative examples in Section \ref{baseline}.

The key designation of divergences is to measure differences between two probability distributions. These distributions can relate to discrete random variables such as symbols in communication systems, or continuous random variables when comparing, for example, two density functions. The differences can also stem from comparing an empirical distribution of some data, and its parametric model. In decision making applications the two distributions could be  aposteriori class probability functions of observations to be classified.  The nuances of these different applications call for divergences of different properties and the  existing spectrum of divergence measures bears witness to the endeavors in the field reported over the decades.

In this paper we focus on the use of divergence to measure incongruence of two classifiers. The problem arises in 
complex decision making systems which often perform sensor data classification tasks using multiple classifiers. Examples of such systems include classifiers processing different modalities of data, ensemble of classifiers aiming to improve classification performance, or hierarchical classification systems where the base classifiers at one level feed their outputs to a contextual classification level. At this decision level the context provided by neighbouring objects is used to improve performance, or derive structural interpretation of the input data. These multiple classifiers voice their opinions about a given set of hypotheses, expressed in terms of aposteriori class probability for each possible outcome.   

In decision making systems engaging multiple classifiers, one would normally expect all the classifiers to support the same hypothesis. A classifier disagreement usually signifies something abnormal; a subsystem malfunction, a sensor data modality being absent, or some anomalous event or situation in the observed  scene. It is  therefore desirable to monitor classifier outputs with the aim of detecting `surprising' classifier incongruence as a trigger for a deeper investigation of its possible causes.

In information theory the magnitude of surprise is intimately linked to the probabilities of the outcome of an experiment. In the decision making context considered in this paper the experiment outcome is the true class membership of a given observation (i.e. finding out which class hypothesis is correct). For outcomes of low probability the surprise is huge, whereas for events that are certain (with probability approaching unity) the surprise is null. The conventional way of measuring the amount of information learnt from an outcome with probability $P$ is using the logarithm of the inverse of $P$. The information gain from an experiment is then measured by averaging over all the possible outcomes.

In the case of classifier incongruence we are interested in measuring the information gain from an outcome involving two or more classifiers. 
For the sake of simplicity, in this paper we shall consider two classifiers only. More specifically, we have two random variables representing class identities, with their distributions, and the question is whether the classifiers agree in supporting the various class hypotheses, or disagree.  The nature of information gain from an experiment changes to a comparison of the respective probabilities of possible outcomes. Congruent classifiers would have similar probability distributions over classes, whereas for incongruent the distributions would be different.

Measuring the information gain from an experiment involving two classifiers is different from quantifying the gain from learning the outcome involving a single classifier. What matters in the case of two classifiers is their comparison. Even if the information gain associated with an experiment involving a single classifier is huge,  if two classifiers have the same aposteriori class probability distribution, they will be congruent. 

A common criterion used for comparison  of the distributions of two random variables is divergence. The most popular divergence measure is the Kullback-Leibler (K-L) divergence, referred to by Itti \cite{Itti-cvpr05} as {\it Bayesian surprise} measure. It has been used as a measure of classifier incongruence by Weinshall \cite{Weinshall-pami2012}, but it is not ideal for a number of reasons:
\begin{enumerate} 
\item \label{reason1} If the distributions are different, the value of incongruence will depend on the actual class probability distributions, rather than on probability differences only.
\item \label{reason2} Measure is asymmetric, i.e. its value depends on which of the two classifier distributions is used as a reference.
\item \label{reason3} Its values are unbounded, which makes it difficult to set a threshold on congruence.
\item \label{reason4} In multi class problems the nondominant classes contribute to clutter, which makes the divergence very noisy.
\end{enumerate}

Some of the above drawbacks have been addressed by alternative divergence measures discussed in Section \ref{survey}. The symmetrized K-L divergence recovers the symmetry property. The Jensen-Shanon divergence \cite{Lin-1991} is both,  symmetric,  and bounds the range of its values to the interval $[0,1]$. However, neither of these measures address properties \ref{reason1} and \ref{reason4}. In search for more suitable candidates one can consider the general family of f-divergences \cite{Liese-2006}. It includes, the divergences based on the Renyi $\alpha-$entropies \cite{Renyi-1961}, of which the commonly used Shannon entropy - the basis of K-L divergence - is a special case for $\alpha=1$.  Another interesting member is, for instance, the $\alpha$-entropy, for $\alpha=\infty$, defined entirely by the probability of the most likely hypothesis, which is used for decision making by each classifier. This choice would avoid the problem of clutter in \ref{reason4}, but this particular property migrates to the associated $\alpha$-divergence family in an undesirable way by focusing on the maximum ratio of a posteriori probabilities, which can emanate from nondominant hypotheses. This can potentially provide a highly misleading information about classifier incongruence.

In this paper we address the problem of measuring classifier incongruence by first introducing the mathematical framework and our baseline - the Kullback-Leibler divergence. This classical information theory divergence is critically assessed in the context of classifier incongruence detection. The critical analysis allows us to identify the properties that a divergence should possess to be able to serve as a measure of classifier incongruence effectively. A brief overview of the options offered by existing tools, and  their ability to satisfy the incongruence measure properties identified provides the motivation for a new measure, called {\it Delta divergence}. Its basis is total variation distance, but we eliminate the clutter by noting that classifier congruence assessment involves only at most three outcomes of material interest: the two classes predicted by the two classifiers, plus the possibility that the true class is neither of the two. The proposed divergence is a function of the absolute value of the difference of the a posteriori class probabilities estimated by the respective classifiers for the  dominant hypotheses. It is shown to  exhibit all the required properties, i.e. being bounded, symmetric, decision cognizant, and decision confidence independent.  The relationship of the proposed divergence with state-of-the-art classifier incongruence measures highlight its advantages which are also confirmed experimentally by showing the effect of clutter on the Kullback-Leibler divergence, as well as on other baseline measures. 

The paper is organised as follows. The related literature is briefly reviewed in Section \ref{survey}. Section \ref{baseline} introduces the mathematical framework and analyses the properties of KL divergence from the point of view of detecting classifier incongruence. As an outcome of this analysis the properties required by any measure of classifier incongruence are postulated in Section \ref{notion}. After a brief discussion of  the properties of other existing tools for measuring classifier incongruence a new divergence is proposed in Section \ref{new} and its properties established in Section \ref{properties}. The novel, {\it decision cognizant} formulation of the classifier incongruence detection problem mitigates the clutter generated by nondominant class hypotheses. This is first shown analytically and later demonstrated experimentally in Section \ref{properties}.  In Section \ref{relations} we discuss the relationship of the proposed divergence with some baseline criteria as well as with the recently advocated heuristic measures of classifier incongruence. Section \ref{conclusions} draws the paper to conclusions.


\section{Related work}
\label{survey}

The introduction of the concept of divergence  is attributed to Jeffreys  \cite{Jeffreys-1946} who proposed it as a measure for comparing the likelihood of two competing hypotheses in statistical hypothesis testing. Jeffrey's divergence is defined as the difference between the means of the log likelihood ratio computed respectively under the two hypotheses.  However, earlier references to the notion of divergence can be traced back to Mahalanobis \cite{Mahalanobis-1930} in his work on measures for comparing two statistical populations, and Bhattacharyya \cite{Bhattacharyya-1943} who proposed to measure the distance between two distributions using the cosine of the angle between the vectors whose components are constituted by the square root of the values of the associated two probability distributions. The Bhattacharyya coefficient is closely related to the Hellinger distance (see e.g. in \cite{Osterreicher-2002}) which dates as far back as 1909.

In spite of the above credits, the key impetus of the intensive study of the topic  over the last six decades was the information theoretic notion of divergence proposed by Kullback and Leibler \cite{Kullback-1951}. Inspired by the seminal work of Shannon \cite{Shannon-1948} on information theory, Kullback and Leibler conceived divergence as the relative gain in information received from an experiment involving two probability distributions relating to the same random variable. 

In their original paper the authors define divergence as the mean information for discrimination between two competing hypotheses. They point out a link between divergence and Fisher's information \cite{Fisher-1925}, and therefore the relevance of the information theoretic notion of divergence to statistical estimation theory. The paper also establishes basic properties of K-L divergence, including its nonnegativity and the conditions that would need to be satisfied for divergence to exhibit the property of transformation invariance. 

One of the factors constraining the use of the K-L divergence involving probability densities is the requirement that the probability distributions are absolutely continuous. 
To overcome this problem, Lin \cite{Lin-1991} proposed an alternative, the Jensen-Shannon divergence, which mitigates this problem and renders his measure more generally applicable.

The information theoretic framework inspired immense interest in theoretical properties of K-L divergence, and led to its generalisation using other entropy functions such as $\alpha$-entropy of Renyi \cite{Renyi-1961}, which includes the K-L divergence as a special case. An even broader generalisation was proposed by Csisz\'ar  \cite{Csiszar-1967} under the name of f-divergences. The family of f-divergences is defined by various choices of convex functions of the likelihood ratio of the respective probability distribution values associated with the alternative hypotheses \cite{Osterreicher-2002}, \cite{Liese-2006}. The family is included in the class of yet more general divergences known as Bregman divergences, see e.g. \cite{Stummer-2012}.

The properties of the numerous divergences have been intensively studied by many authors \cite{Csiszar-1967,Pardo-2003,Liese-2006}. The studies investigate  divergence measure characteristics such as boundedness, finiteness, additivity for independent observations, behaviour under transformation \cite{Pardo-1997}, symmetry, sensitivity to outliers, treatment of inliers, uniqueness, range, behaviour in the case of the two distributions being orthogonal \cite{Frydlova-2012}, convergence of quantised divergences  \cite{Harremoes-2008}, relationships between divergences and their mutual bounds. For instance some divergence measures are less amenable to analytical simplification, and mutual bounds are useful to compare them with those measures that can be analytically developed for certain types of distributions (e.g. K-L divergence for normal distributions). There is interest in establishing the 
existence of metric properties, as well as topological and geometric properties. The study of topological and geometric properties of f-divergences  by Csisz\'ar in \cite{Csiszar-1967a,Csiszar-1975} led to the advocation of perimeter divergences \cite{Osterreicher-1996} and their generalisation proposed by \"Osterreicher and Vajda \cite{Osterreicher-2003}. 

An interesting overview of the properties of f-divergences is presented in \cite{Liese-2006}. The authors provide elegant derivations of the well known properties based on the Taylor expansion of f-divergences, rather than by resorting to the commonly adopted approach based on Jensen's inequality. The subject of properties of divergence measures continues to  generate interest even now, especially in the context of specific applications \cite{Sason-2015}.

In communication systems, divergence is used to measure, for example, communication channel distortion rates and to optimise channel and source coding (see e.g. \cite{Sason-2015}). Similarly, divergences play a role in optimising the quality of audio and video material compression for storage and archival purposes. 

Information-theoretic divergences have application not only in communication systems, but many diverse areas. In statistics, divergence measures have been used for the analysis of contingency tables \cite{Gokhale-1978} and  for estimating the parameters of model distributions \cite{Jeffreys-1946}, gauging the consistency of observations with a hypothesised probability distribution model \cite{Frydlova-2012}, and comparing true distributions with their approximations \cite{Chow-1968}, as well as for comparing stochastic processes over time using divergence rates \cite{Girardin-2015}. 

In the context of statistical decision making, Kailath \cite{Kailath-1967} investigated the relative merits of divergence and Bhattacharyya distance as surrogate criteria for error probability in signal selection for signal detection. In a similar vein, Boekee \cite{Boekee-1979} studied divergence as a criterion for feature selection in pattern recognition and Toussaint \cite{Toussaint-1978} advocated its use instead of error probability for pattern classification. The use of divergence instead of classification error probability may have computational advantages. Most of all, the results in the literature are normally applicable to two class pattern recognition problems only, but some of the divergences, such as the Jensen-Shannon divergence \cite{Lin-1991} support extension to multiclass cases, including error bounds.  Bregman divergences have also been used for non supervised pattern classification and for data analysis based on clustering \cite{Banerjee-2005}.

In this paper, our focus is on the application of divergences for detecting classifier incongruence. Closest to this particular  interest is the use of K-L divergence for gauging classifier incongruence by Weinshall et al \cite{Weinshall-pami2012}. They adopted K-L divergence following Itti and Baldi \cite{Itti-cvpr05} who used it as an objective measure of surprise experienced by subjects reacting to a stimulus induced by the content of a test video. In their experiments divergence was used to compare prior belief captured in terms of a prior distribution, with a new stimulus represented by a posterior distribution. They referred to the K-L divergence in this context as `Bayesian surprise' measure. Some of the deficiencies of the K-L divergence as a measure of classifier incongruence were addressed by heuristic measures proposed in \cite{Kittler-pami2013,Kittler-2015}. In the next section we provide a more principled basis for classifier incongruence detection and develop a novel measure, referred to as Delta divergence, which satisfies the set of desirable properties identified for this specific application.

\section{Delta divergence}
\label{delta}

We start the discussion by introducing the necessary mathematical notation. We then revisit the classical K-L divergence to establish a baseline and to point out some of its deficiencies from the point of view of measuring classifier incongruence. This will allow us to define the notion of classifier incongruence and postulate  the properties a divergence measure should possess to support this particular application. We then consider the spectrum of available divergences to identify a suitable candidate and develop it to a novel classifier incongruence measure that is classifier decision cognizant and reflects the specified properties.

\subsection{Baseline}
\label{baseline}

Let us consider a pattern recognition problem where the object or phenomenon to be recognised is represented by a pattern vector ${\bf x}$ belonging to one of mutually exclusive classes $\omega_i, i=1,....,m$.  Given observation ${\bf x}$, we shall denote the aposteriori probability of its membership in class $\omega_i$ as  $P(\omega_i|{\bf x})$. The automatic assignment of pattern vector ${\bf x}$ to one of the classes is carried out by a classifier employing an appropriate decision function. Regardless of the type of machine learning solution, we shall assume that the classifier effectively computes the aposteriori class probabilities $P(\omega_i|{\bf x}), \forall i $ and engages a Bayesian decision rule to effect the class assignment. 

Let us assume that for the same object or phenomenon there is another classifier which is basing its opinion about the object's class membership on its set of aposteriori class probabilities $\tilde P(\omega_i|{\bf y}), \forall i$, this time based on observation ${\bf y}$. The observation could be the same as ${\bf x}$ but, in general, ${\bf y}$ can be distinct. We are concerned with the problem of measuring the congruence of these two classifiers in supporting the respective hypotheses given the observations ${\bf x}$ and ${\bf y}$. In essence we have two probability distributions, and the classifiers would be deemed congruent if the two probability distributions agree, and incongruent, if the two probability distributions are different. For the sake of simplicity and notational clarity, in the following we shall focus on a specific instance ${\bf x},{\bf y}$ and drop referring to these observations explicitly, using a shorthand notation for the class probabilities as $P_i$ and $\tilde P_i$, i.e.
\begin{equation}
\label{m1}
\begin{array}{lll}
P_i=P(\omega_i|{\bf x}) & \tilde P_i=\tilde P(\omega_i|{\bf y}) & \forall i \\
\end{array}
\end{equation} 

The basic concept in information theory is the notion of self-information. It conveys the amount of information we gain by observing an event $\omega$ which occurs with probability $P(\omega)$. If the probability of occurrence is high, i.e. close to one, we learn very little when the event occurs. However, when the probability $P(\omega)$ is low, the amount of information we gain is huge. Accordingly, self-information $I(\omega)$ is defined as
\begin{equation}
\label{inf1}
 I(\omega) =-\log P(\omega)
\end{equation} 
which takes values from the interval $[0,\infty]$. $I(\omega)$ is referred to as `surprisal', as it quantifies the surprise of seeing a particular outcome. 

In general, when an experiment has a number of possible outcomes $\omega_i, i=1,...,m$, the uncertainty associated with the experiment is expressed in terms of the average information gain from observing the outcome. Let $P_i$ be the probability distribution over the events $\omega_i$. The information gain $h(P)$ is defined as 
\begin{equation}
\label{inf2}
h(P) = -\sum_{i=1}^m P_i \log P_i
\end{equation}
$h(P)$ is known as entropy. It is interesting to note that, as a result of the averaging process, the contribution to entropy made by events with small probability values is low, as
\begin{equation}
\label{inf3}
\lim_{x \rightarrow 0} x \log x =0
\end{equation}

Rather than measuring the information gained from an experiment, here we are interested in assessing the degree of agreement between two probability distributions $P$ and $\tilde P$ estimated over a set of hypotheses $\Omega=\{\omega_i, i=1,...,m\}$ by two different classifiers to gauge whether the classifiers agree in supporting a particular hypothesis or not. This can be achieved by comparing relative uncertainties associated with the two probability distributions $P$ and $\tilde P$. A disagreement in their opinion about the identity of an object being classified would be considered surprising. We therefore need a measure of surprise which compares these two distributions. The classical measure suggested for this purpose is the K-L divergence
\begin{equation}
\label{m6}
D_K=\sum_i \tilde P_i \log \frac{\tilde P_i}{P_i}
\end{equation}
coined Bayesian surprise by Itti  \cite{Itti-cvpr05}, and used for measuring classifier incongruence by Weinshall \cite{Weinshall-pami2012}.


\subsection{Notion of classifier incongruence}
\label{notion}

We know that classifiers compute class aposteriori probabilities to make a decision, and that these probabilities must be involved in the definition of classifier incongruence. However, the notion of classifier incongruence is far from self evident. It is not crisply defined as, for instance, classifier error, or a particular shade of colour. If the class probabilities output by two classifiers are similar, then we would agree that the classifiers are congruent. However, by how much can they differ before they cease to be congruent? If incongruence is like `distance', then the concept is clearly a continuum, rather than a discrete property, and the dichotomy between congruence and incongruence can only be defined by an appropriate threshold. However, what gauge should be used as an incongruence measure?  

To answer these questions and to develop a suitable metric, we shall consider the classical K-L divergence (\ref{m6}) in more detail by elaborating a few special cases that should give us insight regarding the essence of congruence/incongruence. First of all, let us start with the simplest case when all the  aposteriori class probabilities generated by the two classifiers are identical. In such a scenario the K-L divergence $D_K$ will be zero, flagging the status of congruence of the two decision making experts. Next, let us consider the case when the classifiers agree on the dominant hypothesis, $i_{dom}$, and support it with identical strength, i.e. $P_{i_{dom}} =\tilde P_{i_{dom}} $. Clearly, the contribution to the K-L divergence due to the dominant class would be zero. We would probably all agree that in such situation the classifiers would be congruent. Yet the support for the nondominant hypotheses, which we shall refer to as clutter,  given by the two classifiers 
\begin{equation}
\label{inf20}
D_K=\sum_{i, i\neq i_{dom}} \tilde P_i \log \frac{\tilde P_i}{P_i}
\end{equation}
could be substantially different from zero, giving potentially a high value to the K-L divergence.  It is apparent, that for a given threshold, the K-L divergence may give rise to false rejections of congruent classifier outputs.

As the next scenario, we shall investigate the case when the two classifier disagree on the dominant hypothesis, but support the nondominant hypotheses in an identical way. Denoting the respective dominant hypotheses by $i_{dom}$ and $\tilde i_{dom}$,  the K-L divergence in this case will be
\begin{equation}
\label{inf21}
D_K= P_{i_{dom}} \log \frac{ P_{i_{dom}}}{\tilde P_{i_{dom}}} + P_{\tilde i_{dom}} \log \frac{ P_{\tilde i_{dom}}}{\tilde P_{\tilde i_{dom}}}  
\end{equation}
Note in (\ref{inf21}) that the value of K-L divergence in this `zero clutter' case will depend on the actual dominant class probabilities, reflecting the surprisal value in the relative information gained.

\begin{figure}[htbp]
\centering 
	\includegraphics[scale=0.26]{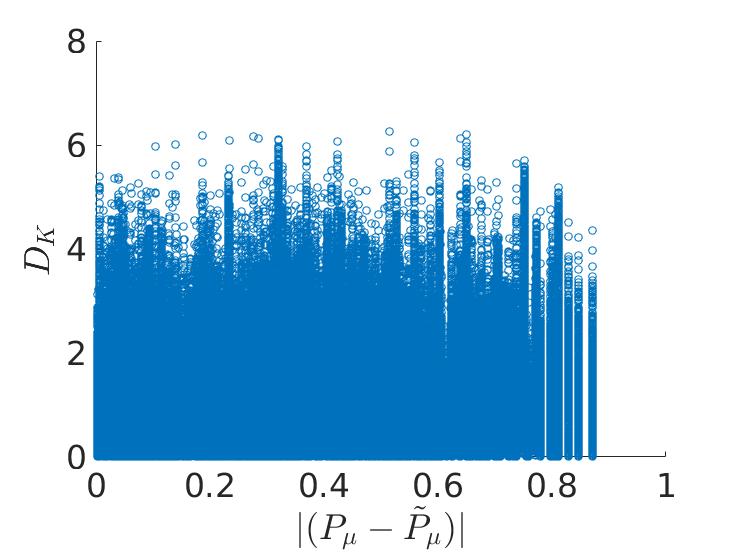}
\caption{Scatter plot of Kullback-Leibler divergence comparing the outputs of two classifiers sampled from a population of probability distributions defined over six classes, constrained by a given difference of posterior class probabilities obtained for the dominant class selected by one of the classifiers.   } \label{KL1}
\end{figure}

\begin{figure}[htbp]
\centering 
	\includegraphics[scale=0.26]{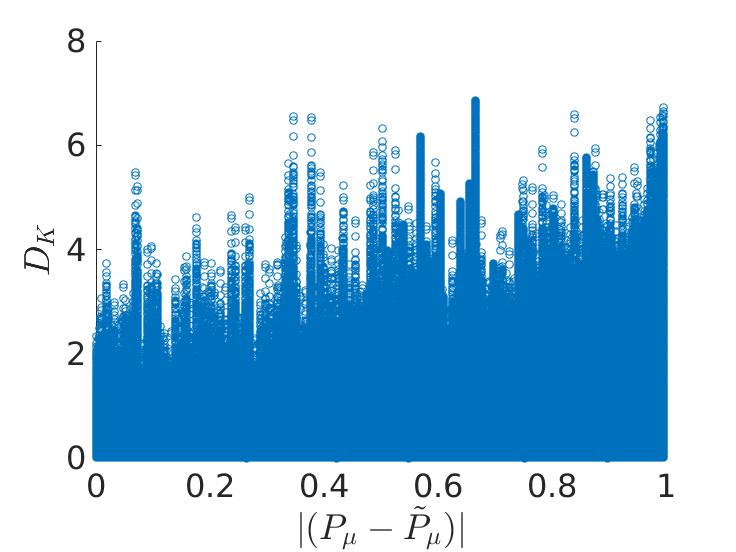}
\caption{Scatter plot of Kullback-Leibler divergence comparing the outputs of two classifiers sampled from a population of probability distributions defined over three classes, constrained by a given difference of posterior class probabilities obtained for the dominant class selected by one of the classifiers.   } \label{KL2}
\end{figure}

Some of these properties are  evident from the scatter plots in Figure \ref{KL1}  and \ref{KL2} which show values of $D_K$ as a function of the difference between the aposteriori probabilities computed by the two classifiers for the dominant hypothesis $\mu$ selected for one of the classifiers.  These values are generated by sampling the space of aposteriori class probability distributions $P_i$ and $ \tilde P_i$, constrained by the choice of $P_{\mu}$ and $\tilde P_{\mu}$. In total one million distributions of $P$ and $\tilde P$ have been drawn for a six class and  a three class case. We can see that for every choice of the difference, the K-L divergence takes values from zero to infinity, even for $|P_{\mu}- \tilde P_{\mu}|=0$. The scatter plot makes it clear that K-L divergence cannot naturally distinguish the state of classifier incongruence from classifier congruence. This is primarily due to the contribution to K-L divergence made by the nondominant hypotheses. We note that the contribution of the nondominant classes becomes more significant for a bigger number of classes as the size of the space of probability distributions increases with $m$. Thus, with increasing number of classes, the false positive rate, i.e. the probability of congruent classifiers being deemed incongruent by $D_K$ will be magnified. At the other end of the spectrum, even for  $|P_{\mu}- \tilde P_{\mu}|=1$, which definitely signifies incongruence, $D_K$ can assume value close to zero. Thus $D_K$ has the capacity to flag a lot of false negative incongruences.

There are a number of conclusions that can be drawn from this analysis. First of all we can see that while `perfect' congruence is independent of the actual values of aposteriori class probabilities of the two distributions, as they are identical, in the case of general congruence and incongruence scenarios, the magnitude of the $D_K$ measure will exhibit strong dependence on the input probability distribution values.   The clutter induced by nondominant classes will create ambiguity, that will degrade the separability of notionally congruent and incongruent classifier cases. It should also be noted that the value of K-L divergence will depend on the class probability distributions used as a reference. If we choose $\tilde P_i$ instead of $P_i$, the observed incongruence value will be different. This is not a useful property for applications where the notion is conceptually symmetric. Also the values of the incongruence measure should be confined to a bounded interval to facilitate the setting of a suitable threshold to dichotomise congruence and incongruent
cases.

From these observation the following desirable properties of the ideal measure of classifier incongruence are beginning to emerge:
\begin{enumerate}
\item \label{Prop1} Overriding focus on dominant hypotheses
\item \label{Prop2} Independence of surprisal content
\item \label{Prop3} Minimum clutter effect
\item \label{Prop4} Symmetry (independence of the choice of distribution as a reference)
\item \label{Prop5} Bounded range of incongruence measure values
\end{enumerate}
Properties 1 and 3 are linked, and suggest that the required measure should concentrate on the dominant hypotheses, and suppress the effect of nondominant classes. Thus the measure we seek should be decision cognizant. Property 2 suggests that classifier incongruence should be a function of differences in aposteriori class probabilities rather than some function of their respective values. The choice of a divergence measure should exhibit symmetry Property \ref{Prop4} and yield values which are bounded, as specified by Property \ref{Prop5}. In the following subsection we shall identify a suitable starting point and develop a novel divergence measure which satisfies the above postulated properties.

\subsection{Delta divergence measure}
\label{new}

Our aim is to develop a divergence that will have all the above stated properties when used as a classifier incongruence measure: namely boundedness, symmetry, being clutter free, and ideally also of low sensitivity to probability estimation errors. Heuristic attempts at finding incongruence gauging measures satisfying these properties were presented in \cite{Kittler-pami2013} and \cite{Kittler-2015}. The key idea in these two papers is to focus on dominant classes as identified by the two classifiers and ignore all the other hypotheses. More specifically, let $\omega = \arg \max_i P_i$ and $\tilde \omega = \arg \max_i \tilde P_i$. These decision dependent measures are defined in \cite{Kittler-pami2013} and \cite{Kittler-2015} respectively as 
\begin{equation}
\label{m8}
\Delta^*=\frac{1}{2}[|P_{\omega}-\tilde P_{\omega}| + |\tilde P_{\tilde \omega}- P_{\tilde \omega}|]
\end{equation}
and
\begin{equation}
\label{m9} 
\begin{array}{ll}
\Delta_{max}&= \frac{1}{2} \max \{|P_{\omega}-\tilde P_{\omega}| + \delta\{\omega,\tilde \omega\} 
|\tilde P_{\tilde \omega}- \tilde P_{ \omega}|, \\
&
 |\tilde P_{\tilde \omega}-  P_{\tilde \omega}| + \delta\{\omega,\tilde \omega\} 
|P_{\omega} - P_{ \tilde \omega}|\}
\end{array}
\end{equation}
where $\delta\{\omega,\tilde \omega\}$ is defined as 
\begin{equation}
\label{m10}
\delta\{\omega,\tilde \omega\}= \left \{ \begin{array}{lll}
0 & if & \omega=\tilde \omega \\
1 & if & \omega \neq \tilde \omega \\
\end{array}
\right .
\end{equation}

In contrast to these heuristic techniques, our objective is to develop a classifier incongruence measure with a solid theoretical underpinning by demanding that it is a proper divergence.
The appropriate toolbox for measuring incongruence between two discrete probability distributions is the family $(h,\phi)$ of functions 
\begin{equation}
\label{m2}
h[\sum_i \phi(P_i,\tilde P_i)]
\end{equation}
with $h()$ and $\phi(P_i, \tilde P_i)$ being polynomial, logarithmic, polylogarithmic, quasi-polynomial, or quasi-polylogarithmic functions \cite{Girardin-2015}, or convex functions \cite{Csiszar-1967}. This family includes Bregman divergences \cite{Stummer-2012}
\begin{equation}
\label{m3}
D_B=\sum_i [f(P_i) - f(\tilde P_i) - (P_i-\tilde P_i) f'(\tilde P_i)]
\end{equation}
the Cziszar f-divergences \cite{Csiszar-1967} reviewed in \cite{Osterreicher-2002} and \cite{Liese-2006}
\begin{equation}
\label{m4}
D_C= \sum_i P_i f(\frac{\tilde P_i}{P_i})
\end{equation}
and the Renyi divergences \cite{Renyi-1961} parameterised by $\alpha$
\begin{equation}
\label{m5}
D_R = \frac{1}{\alpha-1}\log [\sum_i P_i(\frac{\tilde P_i}{P_i})^{\alpha}]
\end{equation}
For an overview the reader is referred to \cite{Girardin-2015}.

Armed with the toolbox, the key question of interest to us is which member of the family would exhibit the properties that reflect the notion of classifier incongruence discussed in Section \ref{notion}. We already established in Section \ref{notion} that the Kullback-Leibler divergence does not.  The Jensen-Shannon divergence  \cite{Lin-1991}
\begin{equation}
\label{m7}
D_J=\frac{1}{2} \sum_i[P_i\log\frac{2P_i}{P_i+ \tilde P_i} + \tilde P_i \log \frac{\tilde 2P_i}{P_i+ \tilde P_i}]
\end{equation}
confines its values to a bounded interval, and is symmetric. However, the contributions to divergence generated by a difference in probabilities for a particular hypothesis are a function of the probabilities themselves, which does not satisfy Property \ref{Prop2} in Section \ref{notion}. Most importantly, all the measures, including Jensen-Shannon divergence, are affected by the divergence clutter injected by weakly supported hypotheses. This clutter is also likely to aggravate the sensitivity of these divergence measures to noise.

Herein we set to develop an incongruence measure which is a member of the family of  divergences in (\ref{m2}). This is the most general family of divergences which has the potential to source the starting point of our development. We start by choosing 
\begin{equation}
\label{m11}
h(z)=z
\end{equation} 
in (\ref{m2}) and opting for the family of f-divergences in (\ref{m4}). The requirement that our starting point satisfies the property that the contribution to divergence is dependent purely on differences in probabilities, rather than their actual values retains only the {\it total variation distance} from this family of f-divergences, defined as
\begin{equation}
\label{m12}
D_T=\frac{1}{2}\sum_i P_i |\frac{\tilde P_i}{P_i}-1|=\frac{1}{2} \sum_i |\tilde P_i - P_i|
\end{equation}
This measure is symmetric and bounded, taking values from the interval $[0,1]$. 

The measure in (\ref{m12}) is still affected by clutter of nondominant classes. The effect of clutter can significantly be reduced by the following argument: When we compare the outputs of two classifiers, there are only three outcomes of interest: the dominant class $\omega$ identified by the classifier with probability distribution $P$, the dominant class $\tilde \omega$ identified by the other classifier, and neither of the two, in other words $\hat \omega = \Omega - \omega - \tilde \omega$. We thus define a new decision cognizant divergence $D_{\Delta}$, which we name Delta divergence,  as
\begin{equation}
\label{m121}
D_{\Delta}=\frac{1}{2}[\sum_{i\epsilon \{\omega, \tilde \omega\}} |\tilde P_i - P_i| +|\tilde P_{\hat \omega} -
P_{\hat \omega}|]
\end{equation}

Noting that the outcome $\hat \omega$ arises with the complement probabilities we can develop Delta divergence, $D_{\Delta}$, in (\ref{m121}) further by considering the  cases when the labels of the dominant classes identified by the two classifiers agree and when they disagree.

\subsubsection{Label agreement}
When the labels agree, i.e. $\omega=\tilde \omega$, the complement probabilities for the event that the true class is not $\omega$ are $1-\tilde P_{\omega}$ and $1 - P_{\omega}$. Then the Delta divergence  in (\ref{m121})  can be expressed
\begin{equation}
\label{m13}
\begin{array}{ll}
D_{\Delta}& = \frac{1}{2}[|\tilde P_{\omega} - P_{\omega}| + |1- \tilde P_{\omega} - 1 + P_{\omega}| ] = \\
&=|\tilde P_{\omega} - P_{\omega}| \\
\end{array}
\end{equation}
In other words, the classifier incongruence can be measured simply by comparing the probabilities of the dominant hypothesis output by the two classifiers.

\subsubsection{Label disagreement}
When the dominant labels identified by the two classifiers disagree, the probabilities of the event $\hat \omega$ that neither of the two dominant classes is the true class are given as
\begin{equation}
\label{m14}
\begin{array}{c}
P_{\hat \omega} = 1 - P_{\omega} - P_{\tilde \omega} \\
\tilde P_{\hat \omega} = 1 - \tilde P_{\omega} - \tilde P_{\tilde \omega} \\
\end{array}
\end{equation}
In this scenario the Delta divergence becomes
\begin{equation}
\label{m15}
\begin{array}{ll}
D_{\Delta}& =\frac{1}{2}[ |\tilde P_{\tilde \omega} - P_{\tilde \omega}|+ | P_{\omega} - \tilde P_{\omega}| + \\
& +  |\tilde P_{\tilde \omega} - P_{\tilde \omega} + \tilde P_{ \omega} - P_{\omega} | ] \\
& =\frac{1}{2}[|A| + |B| + | A - B| ]\\
\end{array}
\end{equation}
Note that the terms $A$ and $B$ can either be both positive, or one of them positive and the other negative. It can be easily shown that it is impossible for both terms to be negative. Consider, for instance, the case $A <0$, i.e. $\tilde P_{\tilde \omega} - P_{\tilde \omega}<0$. Then, since $ P_{\tilde \omega} < P_{\omega}$ ($\omega$ being the dominant class for classifier with distribution $P$), and $ \tilde P_{\tilde \omega} > \tilde P_{\omega}$ ($ \tilde \omega$ being the dominant class for classifier $\tilde P $ ) we have
\begin{equation}
\label{m151}
0 < P_{\tilde \omega} - \tilde P_{\tilde \omega} <  P_{\omega} - \tilde P_{\omega}
\end{equation}
The positivity of $A$ when $B$ is negative can be shown in the same way.

Now suppose $A$ is negative. Then  $A-B$ in (\ref{m15}) is also negative, and its absolute value is equal $|A+B|=|A|+|B|$. If, on the other hand, $B$ is negative, then $-B$ is positive, and the absolute value of $A-B$ will again equal $|A|+|B|$. Thus when one of the terms, $A$ and $B$ is negative, Delta divergence will be
\begin{equation}
\label{m152}
D_{\Delta}= [|A|+|B|]=[|\tilde P_{\tilde \omega} - P_{\tilde \omega}| +
| P_{\omega} - \tilde P_{\omega}|]
\end{equation}

When both $A$ and $B$ are positive, the term $A-B$ is either positive, or negative, depending on the relationship of $A$ and $B$. If $A>B$, then the difference will be positive and we can ignore the absolute value operation, i.e. $|A-B|=A-B$. If $A<B$, then the difference will be negative and $|A-B|=B-A$. Thus we can write for $D_{\Delta}$ in (\ref{m15})
\begin{equation}
\label{m16}
D_{\Delta}=\left \{ \begin{array}{lll}
A & if & A\ge B \\
B & if & A < B \\
\end{array}
\right .
\end{equation}
\subsubsection{Delta divergence overview}
Combining the results for these scenarios yields a surprisingly simple divergence measure for gauging classifier incongruence, .i.e.

\begin{equation}
\label{m17}
\begin{array}{l}
D_{\Delta}= \\
\left \{ \begin{array}{lll}
|\tilde P_{\omega} - P_{\omega}| & \omega=\tilde \omega & \\
\max \{ |\tilde P_{\tilde \omega} -  P_{\tilde \omega}|, |P_{\omega} - \tilde P_{\omega}| \} & \omega \ne \tilde \omega & \hspace{0.3cm}
A \ge 0, B \ge 0  \\
{[|\tilde P_{\tilde \omega} -  P_{\tilde \omega}| + | P_{\omega} - \tilde P_{\omega} |] } &
\omega \ne \tilde \omega & \left \{ \begin{array}{l}
\hspace{-0.2cm} A<0,B\ge0 \\
\hspace{-0.2cm} A\ge0,B<0 \\
\end{array}
\right . \\
\end{array}
\right .
\\
\end{array}
\end{equation}
In other words, the incongruence measure is defined either by the maximum absolute value difference between the probabilities output by the two classifiers for the respective dominant hypotheses or by the sum of these differences. 

The measure has attractive properties.  It is zero, whenever the aposteriori probabilities for the shared dominant class are identical, regardless of the differences in the distribution of the residual probability mass over all the other classes. As it always involves the difference of two probability values it is symmetric. Also, its sensitivity to estimation errors should be very low. It has a monotonic transition between the function values for the label agreement and label disagreement cases. In fact, as we move from the label agreement to the label disagreement case, when another class for the second classifier begins to assume the dominant role, Delta divergence will continue increasing (potentially by a step change) by virtue of the growing difference between the dominant class probability of the first classifier and the support for this hypothesis voiced by the second classifier.

\subsubsection{Two class case}
\label{twoclass}
In the two class case, when the classifiers agree on the dominant hypothesis $\omega$, Delta divergence is given as
\begin{equation}
\label{m171}
D_{\Delta}= \frac{1}{2}[|P_{\omega}-\tilde P_{\omega}| + |1 - P_{\omega} - 1 + \tilde P_{\omega}| ] = |P_{\omega}-\tilde P_{\omega}|
\end{equation}
In the label disagreement case, the set of nondominant hypotheses is empty. Hence the Delta divergence has just two terms that are identical to those in (\ref{m171}). Thus the general formula for $D_{\Delta}$ in the case of agreement and disagreement is as given in (\ref{m171}).

\subsection{Properties of Delta divergence}
\label{properties}
In this section we briefly review the properties of Delta divergence and verify that it satisfies the characteristics specified in Section \ref{notion}. In addition we shall determine the conditions under which the proposed divergence measure is a metric. This particular property is interesting in the context of assessing incongruence of more than two classifiers.
\begin{itemize}
\item {\bf Decision cognizance property:} The Delta divergence proposed in (\ref{m17}) is defined in terms of the aposteriori class probabilities associated with the dominant hypotheses identified by the two classifiers. The measure therefore focuses only on the dominant class hypotheses as required by property \ref{Prop1} in Section \ref{notion}.
\item {\bf Surprisal independence:} The proposed divergence is defined in terms of differences in aposteriori class probabilities of the dominant hypotheses, rather than their respective values. Thus the value of delta divergence is independent of the base level of these probabilities, and consequently of the surprisal values.
\item {\bf Robustness to clutter:} The advantage of Delta divergence over total variation distance can be demonstrated by comparing the contributions of the nondominant hypotheses to these two measures. In the case of Delta divergence the implicit contribution to `clutter' is given by $\frac{1}{2}|P_{\hat \omega}-\tilde P_{\hat \omega}|$ where $\hat \omega$ represents the set of nondominant classes. In the case of total variation distance the `clutter' contribution becomes
\begin{equation}
\label{m172}
\frac{1}{2}\sum_{
i \epsilon \hat \omega }
|P_i -\tilde P_i|
\end{equation}
Rearranging the clutter contribution to Delta divergence we have
\begin{equation}
\label{m173}
\frac{1}{2}|P_{\hat \omega}-\tilde P_{\hat \omega}| =\frac{1}{2}| \sum_{
i \epsilon \hat \omega 
} [P_i -\tilde P_i]  | 
\leq \frac{1}{2}\sum_{
i \epsilon \hat \omega } 
|P_i -\tilde P_i| \\
\end{equation}
Thus the sensitivity of Delta divergence to clutter is significantly lower than that of total variation distance.

It is interesting to note, that if the first two terms in (\ref{m15}) are considered as `pure incongruence measure' (PIM) and the last term as a group clutter, $D_{\Delta_{clutter}}$, then from (\ref{m17}) we conclude 
\begin{equation}
\label{m174}
D_{\Delta_{clutter}} = \left \{ \begin{array}{ll}
\hspace{-0.2cm} \frac{1}{2}|\tilde P_{\tilde \omega} +  \tilde P_{\omega} -  P_{\omega} - P_{\tilde \omega} |& \left \{
\begin{array}{l}
\hspace{-0.2cm}\tilde P_{\tilde \omega} - P_{\tilde \omega} \ge 0 \\
 \hspace{-0.2cm}P_{\omega} - \tilde P_{\omega} \ge 0 \\
\end{array}
\right . \\
\hspace{-0.2cm} \frac{1}{2} \times PIM & elsewhere \\
\end{array}
\right .
\end{equation}
This shows that the contributed group clutter is equal to the magnitude of pure incongruence measure in most cases. When the labels of the dominant hypotheses selected by the classifiers disagree, and the difference between the probability for the top ranking hypothesis rendered by the supporting classifier relative to the other classifier is nonnegative, the group clutter equals one half of the difference of the two differences.   Alternatively, the clutter is equal to the difference between the support for the union of the two hypotheses. Thus in this particular case the clutter is  proportional to the difference between the residual probability masses associated with the nondominant classes.     

The superiority of Delta divergence $D_{\Delta}$ over $D_T$ from the clutter point of view is also evident from the experimental results shown in Figure \ref{clutter1} and \ref{clutter2}. The figures present the scatter plot of clutter injected in the total variation divergence $D_T$ against the clutter of $D_{\Delta}$, as defined on the right hand side and left hand side of (\ref{m173}) respectively. The values of the clutters are computed by sampling the populations of aposteriori class probability distributions of the two classifiers for three-class and six-class problems as described in Section \ref{notion}. It should be noted that the clutter affecting $D_T$ in the three class case is less severe than in the six class case, because the scope for cluttering in the former case is considerably limited. In the two class case it disappears altogether. By the same token, in pattern recognition problems involving a large number of classes, the induced clutter can dominate the value of total variation divergence and make it impossible to detect classifier incongruence reliably. 

\begin{figure}[htbp]
\centering 
	\includegraphics[scale=0.26]{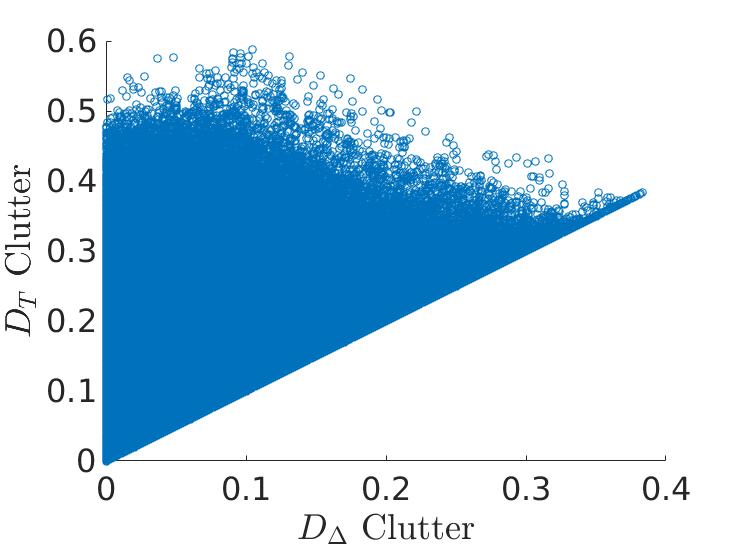}
\caption{Scatter plot of $D_T$ clutter  against $D_{\Delta}$ clutter, affecting the outputs of two classifiers using these two divergences. The values are computed for samples from a population of probability distributions defined over six classes.   } \label{clutter1}
\end{figure}  

\begin{figure}[htbp]
\centering 
	\includegraphics[scale=0.26]{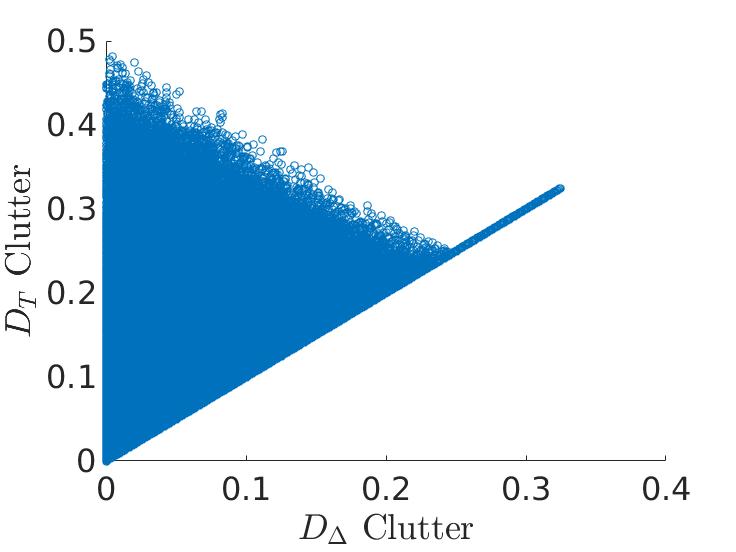}
\caption{Scatter plot of $D_T$ clutter  against $D_{\Delta}$ clutter, affecting the outputs of two classifiers using these two divergences. The values are computed for samples from a population of probability distributions defined over three classes.   } \label{clutter2}
\end{figure}

 \item {\bf Bounded range:} Inspecting (\ref{m17}), it is evident that its values satisfy $0 \leq D_{\Delta} \leq 1$. Hence Delta divergence is bounded to interval $[0,1]$ in compliance with Property \ref{Prop4} of Section \ref{notion}.

\item {\bf Symmetry:} As (\ref{m17}) involves only differences of aposteriori class probabilities, $D_{\Delta}$ is symmetric in compliance with property \ref{Prop5} of Section \ref{notion}.
\item {\bf Metric property:} The total variation distance, $D_T$, from which the proposed divergence has been developed is a metric. This can easily be checked by considering three classifiers $A,B,C$ with probability distributions $P$, $\tilde P$ and $\hat P$ respectively. The sum of variation distances $D_{AB}$ and $D_{BC}$ can be written as 
\begin{equation}
\label{m19}
\begin{array}{l}
D_{AB}+D_{BC}=\sum_i [|P_i-\tilde P_i| + | \tilde P_i - \hat P_i|] \geq \\
\geq \sum_i [|P_i-\tilde P_i +  \tilde P_i - \hat P_i|] = \\
=\sum_i [|P_i - \hat P_i|] =  D_{AC}\\
\end{array}
\end{equation} 
The metric property does not extend to $D_{\Delta}$ because of the clutter reducing operation of merging all nondominant hypotheses into a single event, as the resulting sets for the three classifiers can be different. However, in the two class  case when the set of nondominant hypotheses is empty, the Delta divergence (\ref{m121}) will degenerate to the total variation distance (\ref{m12}), and the incongruence measure will become a metric.  
\end{itemize}

\section{Relationship of $D_{\Delta}$ to other measures}
\label{relations}

As we have developed Delta divergence from the total variation distance it is pertinent to elaborate the key differences between these two divergences. The main distinguishing feature of Delta divergence is the way it deals with clutter. Let us denote by $\Omega^+$ the set of dominant hypotheses identified by the two classifiers, which will have a single element for label agreement and two elements for label disagreement. The complement set $\Omega^-$ is constituted by all the nondominant hypotheses, i.e. $\Omega^- = \Omega-\Omega^+$ and the probability of one of its members being the true class is $P_{\Omega^-} = \sum_{i \epsilon \Omega^-} P_i$  and $\tilde P_{\Omega^-} = \sum_{i \epsilon \Omega^-} \tilde P_i$ respectively for the two classifiers. Referring to (\ref{m12}), we can express $D_{\Delta}$ as
\begin{equation}
\label{m131}
\begin{array}{ll}
D_{\Delta}& = \frac{1}{2}[\sum_{i \epsilon \Omega+} |P_i- \tilde P_i| + |P_{\Omega^-} - \tilde P_{\Omega^-}|  ]  \leq \\
& \leq \frac{1}{2}[\sum_{i \epsilon \Omega+} |P_i- \tilde P_i| +  \sum_{i \epsilon \Omega^-} |P_i- \tilde P_i|] = D_T \\
\end{array}
\end{equation} 
Thus $D_{\Delta} \leq D_T$, with equality only for the two class case $m=2$. Even for $m=3$ the total variation distance will be greater than Delta divergence because the set of nondominant hypotheses will contain more than one element in the case of label agreement. 

\begin{figure}[htbp]
\centering 
	\includegraphics[scale=0.13]{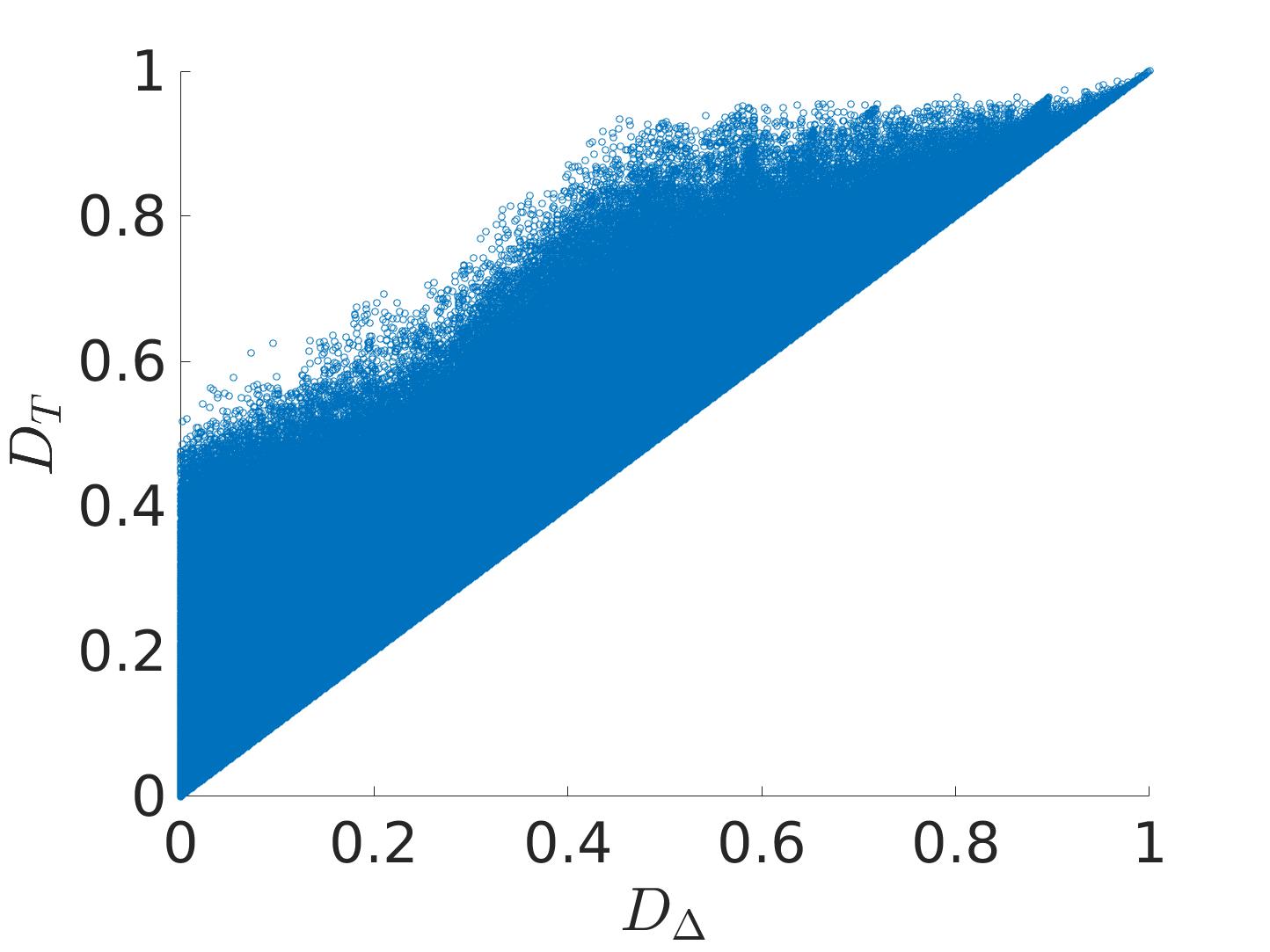}
\caption{Scatter plot of total variation divergence $D_T$ against Delta divergence $D_{\Delta}$, comparing the outputs of two classifiers sampled from a population of probability distributions defined over six classes.   } \label{total1}
\end{figure}

\begin{figure}[htbp]
\centering 
	\includegraphics[scale=0.13]{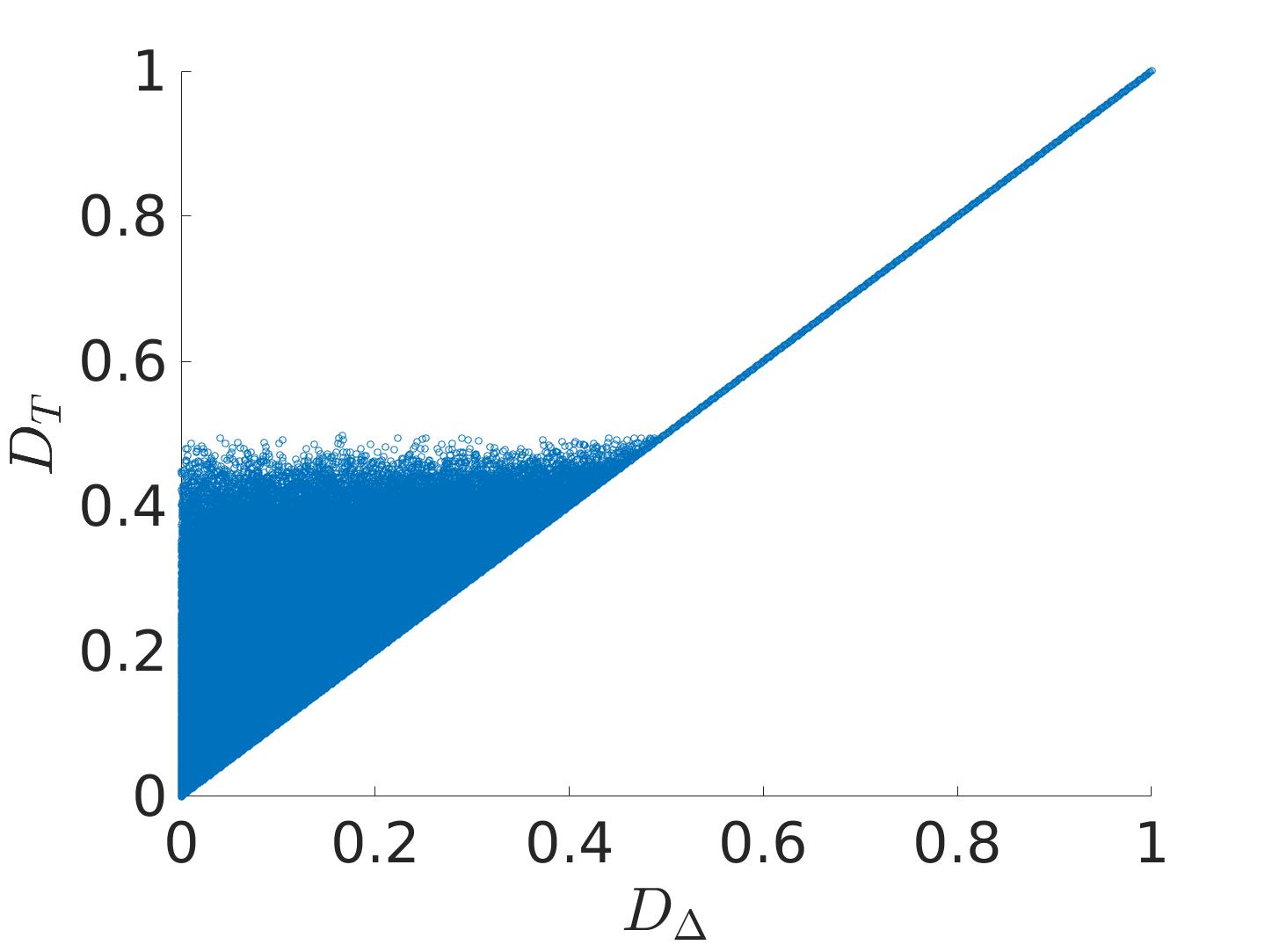}
\caption{Scatter plot of total variation divergence $D_T$ against Delta divergence $D_{\Delta}$, comparing the outputs of two classifiers sampled from a population of probability distributions defined over three classes.   } \label{total2}
\end{figure}

The relationship between these two divergences is shown in Figures \ref{total1}  and \ref{total2} which plot values of $D_T$ against $D_{\Delta}$. It should be noted that for every value of $D_{\Delta}$ there are many possible values of $D_T$, as already shown in Section \ref{properties}. These have been identified by sampling the probability distributions $P$ and $\tilde P$ for a fixed value of $D_{\Delta}$, as described in Section \ref{notion}, to obtain the scatter plots in Figure \ref{total1} and \ref{total2}. It is apparent from the plots that for $m=6$ the distribution scenarios are much less heavily constrained than for $m=3$, resulting in much greater differences in the values of $D_T$ and $D_{\Delta}$. Let us consider any case with $m\geq3$, and a given threshold, $D_{\Delta_T}$, dichotomising congruence and incongruence. It is apparent that for any $D_{\Delta} \leq D_{\Delta_T}$ the total variation divergence $D_T$ can take almost any value from the $[0,1]$ range, which means that due to clutter it will be impossible to discriminate  between congruent and incongruent cases.

\begin{figure}[htbp]
\centering 
	\includegraphics[scale=0.13]{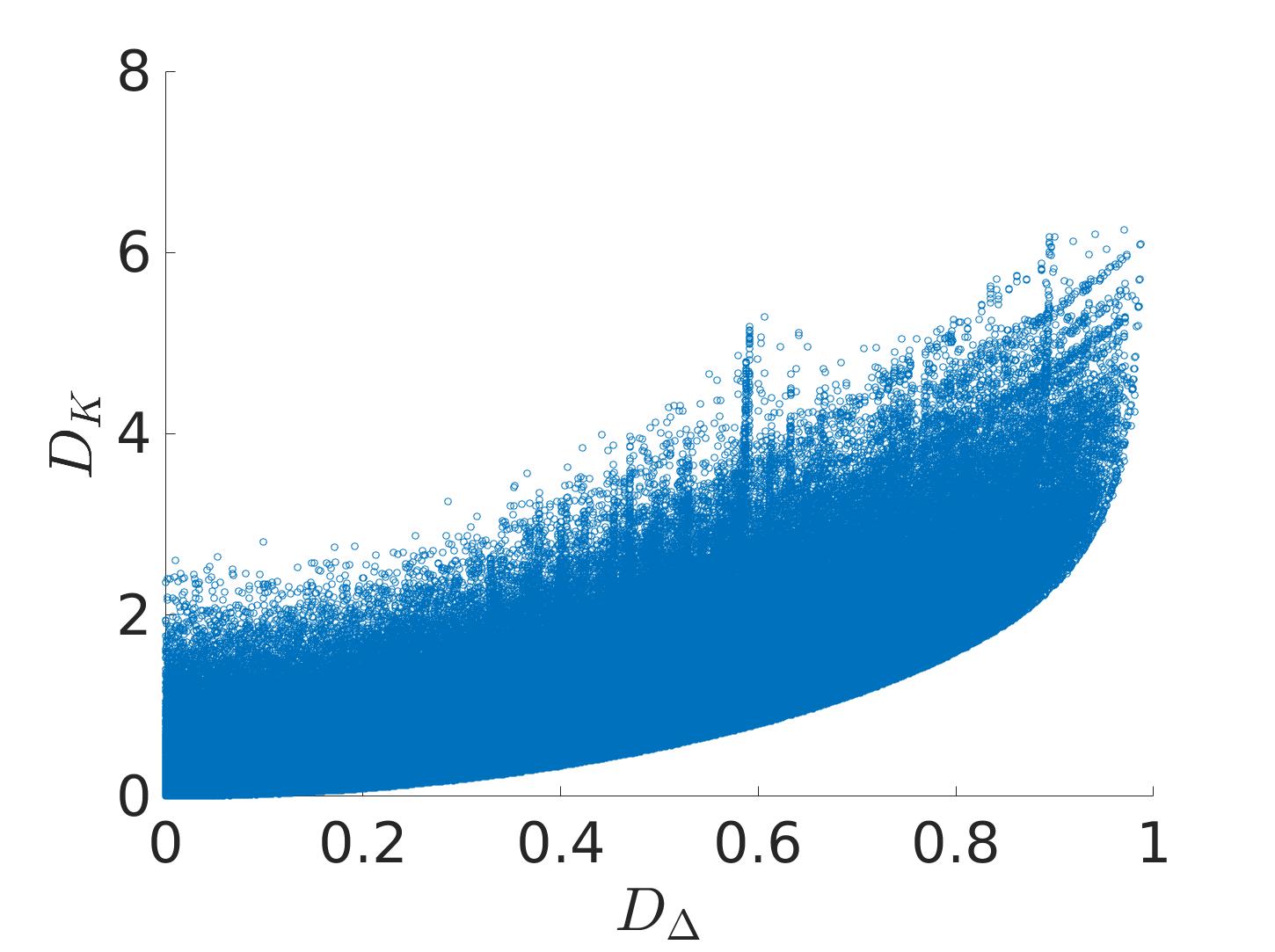}
\caption{Scatter plot of Kullback Leibler divergence $D_K$ against Delta divergence $D_{\Delta}$, comparing the outputs of two classifiers sampled from a population of probability distributions defined over six classes.   } \label{Bayesian1}
\end{figure}

\begin{figure}[htbp]
\centering 
	\includegraphics[scale=0.13]{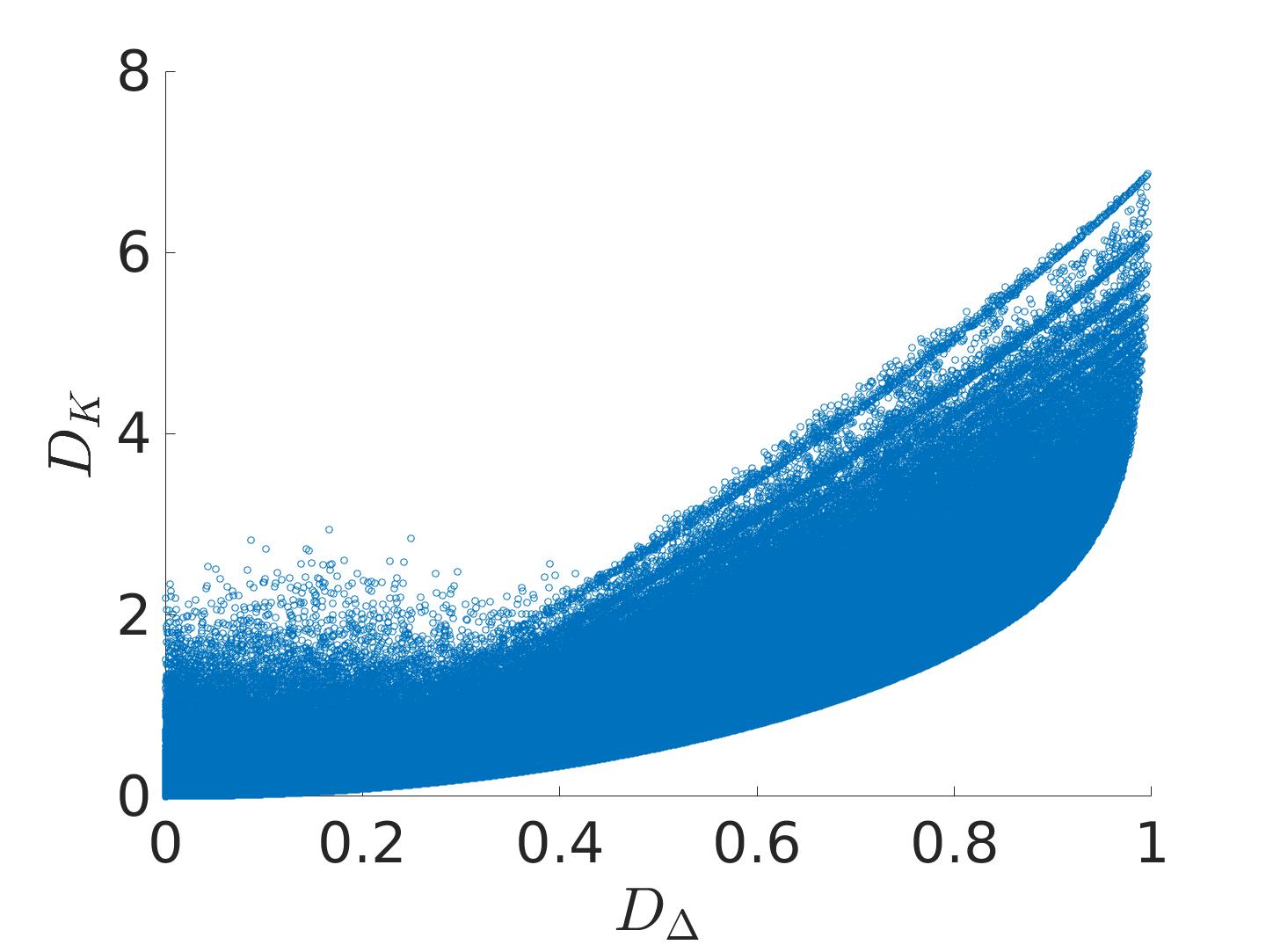}
\caption{Scatter plot of  Kullback Leibler divergence $D_K$ against Delta divergence $D_{\Delta}$, comparing the outputs of two classifiers sampled from a population of probability distributions defined over three classes.   } \label{Bayesian2}
\end{figure}

Next we compare Delta divergence with K-L divergence. The same experiment, involving the sampling of the space of probability distributions $P$ and $\tilde P$ for fixed values of $D_{\Delta}$, is conducted and the results plotted in Figures \ref{Bayesian1} and \ref{Bayesian2} for a six class and three class problems. We note a number of observations: First of all the range of values assumed by the K-L divergence is much greater, which would make it difficult to set a suitable threshold between classifier congruence and incongruence. The unbounded range reflects the dependence of K-L divergence on the surprisal values of the additive terms in the expression for the K-L divergence. In the plots the observed values greater than $D_K=8$ are not shown to avoid undue data compression. We also note that the greater the number of classes, the greater the variation of K-L divergence values caused by the contribution of the clutter of nondominant classes. The clutter is responsible for a significant overlap of K-L divergence values for the classifier congruence and classifier incongruence cases. This can be seen by drawing horizontal lines cutting the scatter plots at different K-L divergence thresholds and noting the resulting distributions (data scatters). For instance setting the threshold to $D_K=3$ will retain many cases with a high value of $D_{\Delta}$ in the congruent category, leading to underdetection of incongruence. Lowering the threshold to, say, $0.75$ will miss many cases with low value of Delta divergence, resulting in a high proportion of false positives.   

It is also interesting to compare Delta divergence with the heuristic measures of classifier incongruence in (\ref{m8}) and (\ref{m9}) which are based on intuition, rather than information theoretic foundations. It can be easily verified that $\Delta^*$ in (\ref{m8}) is exactly the same as Delta divergence in (\ref{m17}) in all cases with the exception of positive values of $A$ and $B$ in label disagreement. In other words we can can express $D_{\Delta}$ as
\begin{equation}
\label{m181}
\begin{array}{l}
D_{\Delta}= \\
\left \{ \begin{array}{lll}
D^* & \omega = \tilde \omega & \\
\max \{ |\tilde P_{\tilde \omega} -  P_{\tilde \omega}|, |P_{\omega} -  \tilde P_{\omega}|\}  &
 \omega \ne \tilde \omega & \left \{ \begin{array}{l}
\hspace{-0.2cm} \tilde P_{\tilde \omega} - P_{\tilde \omega} \ge 0 \\
\hspace{-0.2cm} P_{\omega} - \tilde P_{\omega} \ge 0 \\
\end{array}
\right . \\
2D^* & \begin{array}{l} 
else-\\
where \\
\end{array}
 & \\
\end{array}
\right .
\\
\end{array}
\end{equation}

In the case of label agreement also (\ref{m9}) equals one half of Delta divergence. In the case of label disagreement, a number of situations may arise. When all the terms in $\Delta_{max}$ are positive, we can drop the absolute value operations and after rearrangement $\Delta_{max}$ becomes 
\begin{equation}
\label{m18}
\begin{array}{ll}
\Delta_{max} &  = \frac{1}{2} \max \{ P_{\omega}+\tilde P_{\tilde \omega} - 2\tilde P_{\omega}, 
P_{\omega}+\tilde P_{\tilde \omega} - 2 P_{\tilde \omega} \} = \\
  &  = \frac{1}{2}[P_{\omega}+\tilde P_{\tilde \omega}]-
\min \{ \tilde P_{\omega}, P_{\tilde \omega} \} \\
\end{array}
\end{equation}
When either $P_{\omega}- \tilde P_{\omega} < 0$ or $\tilde P_{\tilde \omega}-  P_{\tilde \omega} < 0$ it can easily be  shown that
\begin{equation}
\label{m181}
\Delta_{max}  = \left\{ \begin{array}{ll}
\frac{1}{2}[P_{\omega}+\tilde P_{\tilde \omega}]- P_{\tilde \omega} & P_{\omega}- \tilde P_{\omega} < 0 \\
\frac{1}{2}[P_{\omega}+\tilde P_{\tilde \omega}]- \tilde P_{\omega}  & \tilde P_{\tilde \omega}-  P_{\tilde \omega} < 0\\
\end{array}
\right.
\end{equation}

Thus in contrast to Delta divergence, in the case of label disagreement $\Delta_{max}$ picks as incongruence value the average of the probabilities associated with the top hypotheses of the two classifiers minus a term corresponding to one of the probabilities assigned by each classifier to the dominant class identified by the other classifier.  
The key point to note is that neither of the heuristic measures, $\Delta^*$ and $\Delta_{max}$, is a  member of the family of divergence functions.    

\begin{figure}[htbp]
\centering 
	\includegraphics[scale=0.13]{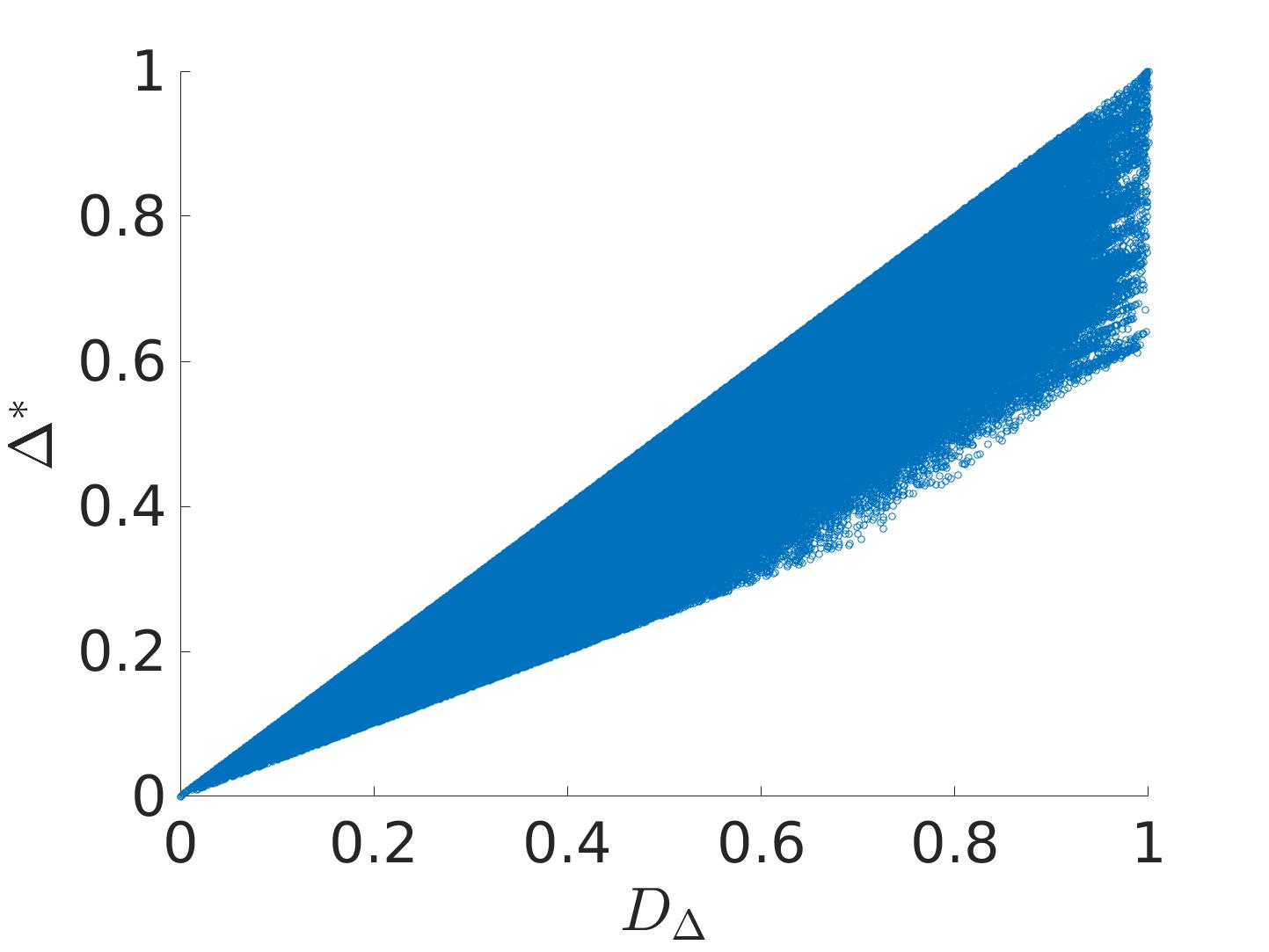}
\caption{Scatter plot of $\Delta^*$   against Delta divergence $D_{\Delta}$, comparing the outputs of two classifiers sampled from a population of probability distributions defined over six classes.   } \label{heuristic21}
\end{figure}

\begin{figure}[htbp]
\centering 
	\includegraphics[scale=0.13]{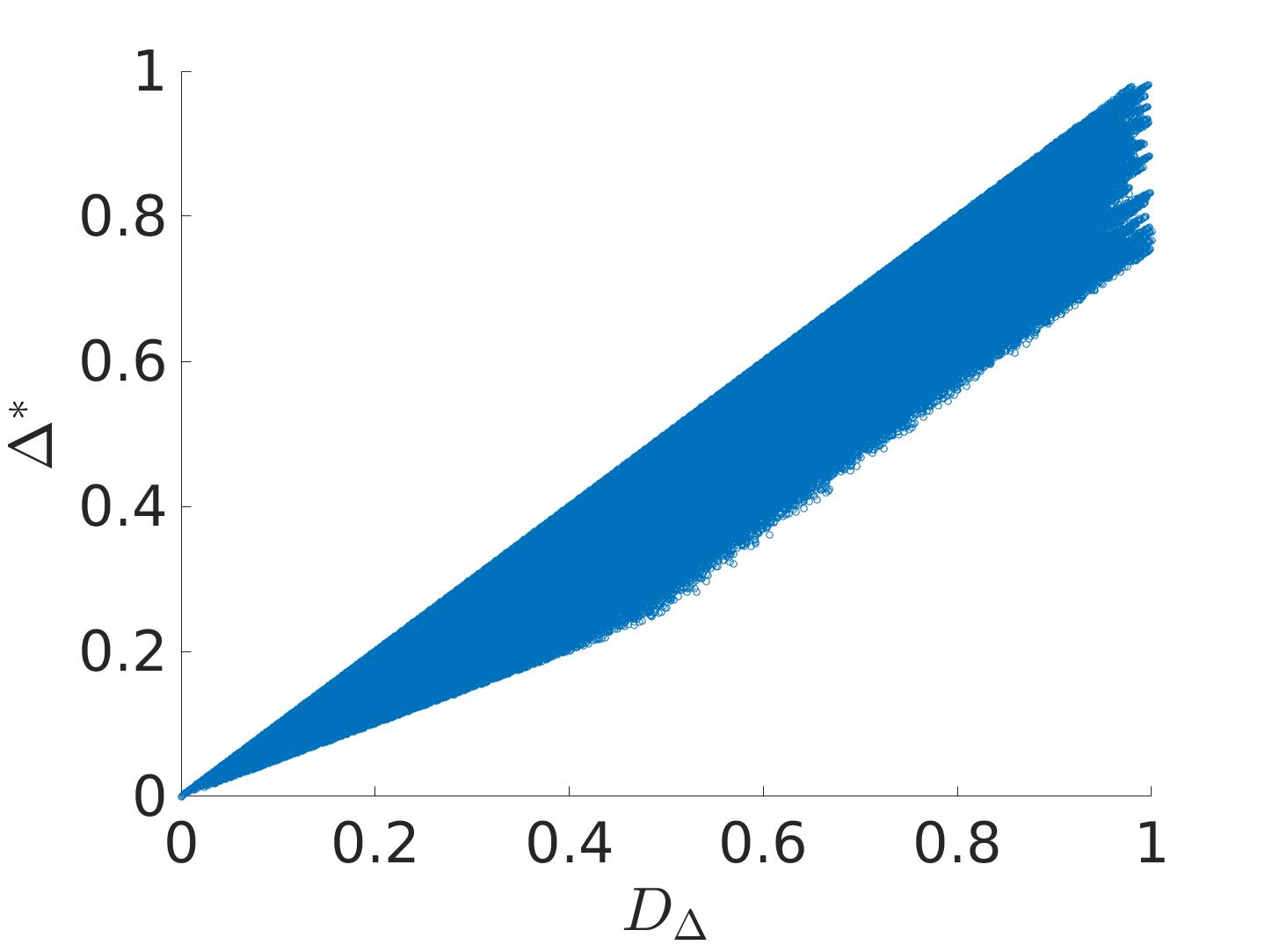}
\caption{Scatter plot of $\Delta^*$  against Delta divergence $D_{\Delta}$, comparing the outputs of two classifiers sampled from a population of probability distributions defined over three classes.   } \label{heuristic22}
\end{figure}

\begin{figure}[htbp]
\centering 
	\includegraphics[scale=0.13]{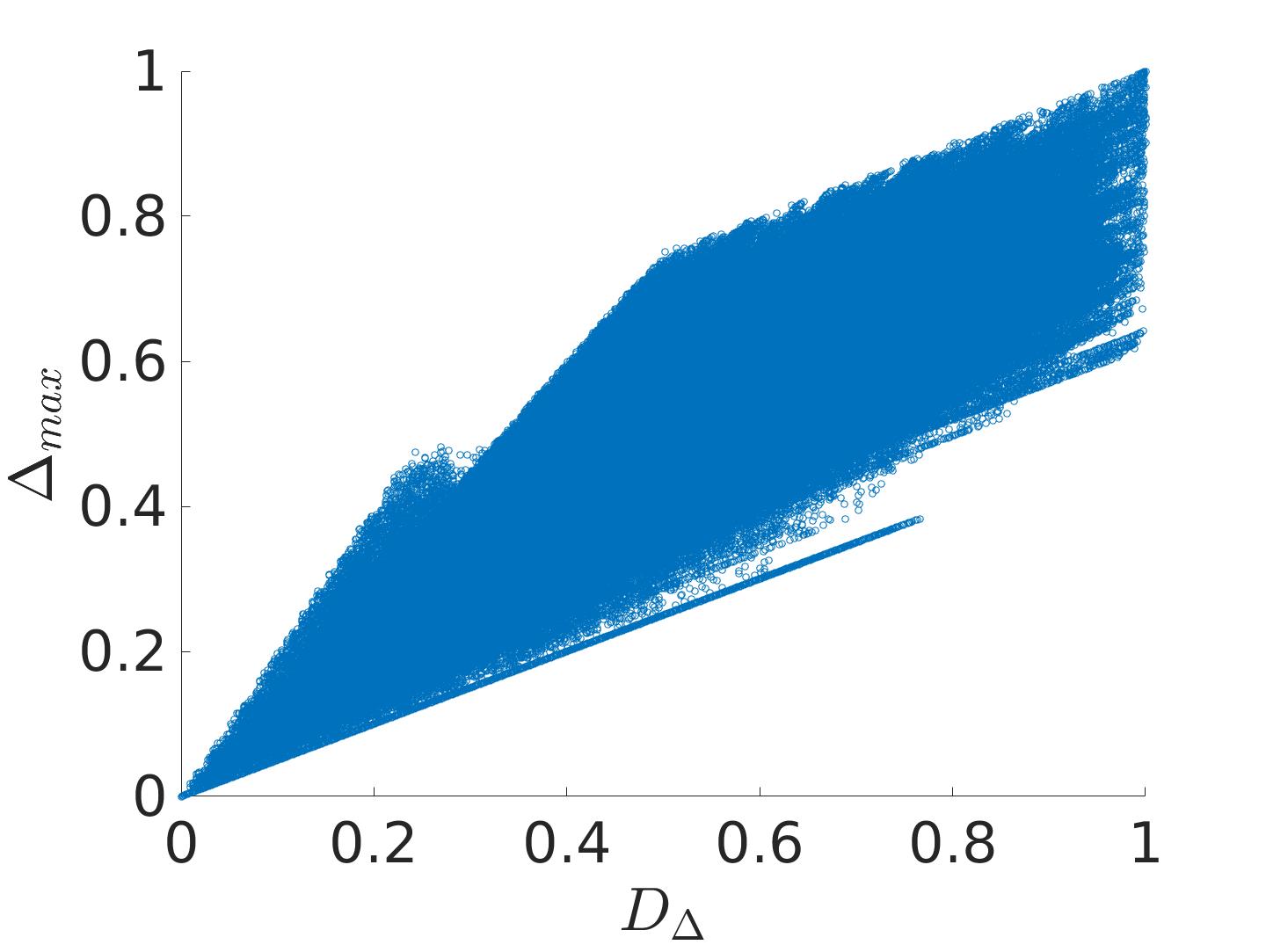}
\caption{Scatter plot of $\Delta_{max}$  against Delta divergence $D_{\Delta}$, comparing the outputs of two classifiers sampled from a population of probability distributions defined over six classes.   } \label{heuristic11}
\end{figure}

\begin{figure}[htbp]
\centering 
	\includegraphics[scale=0.13]{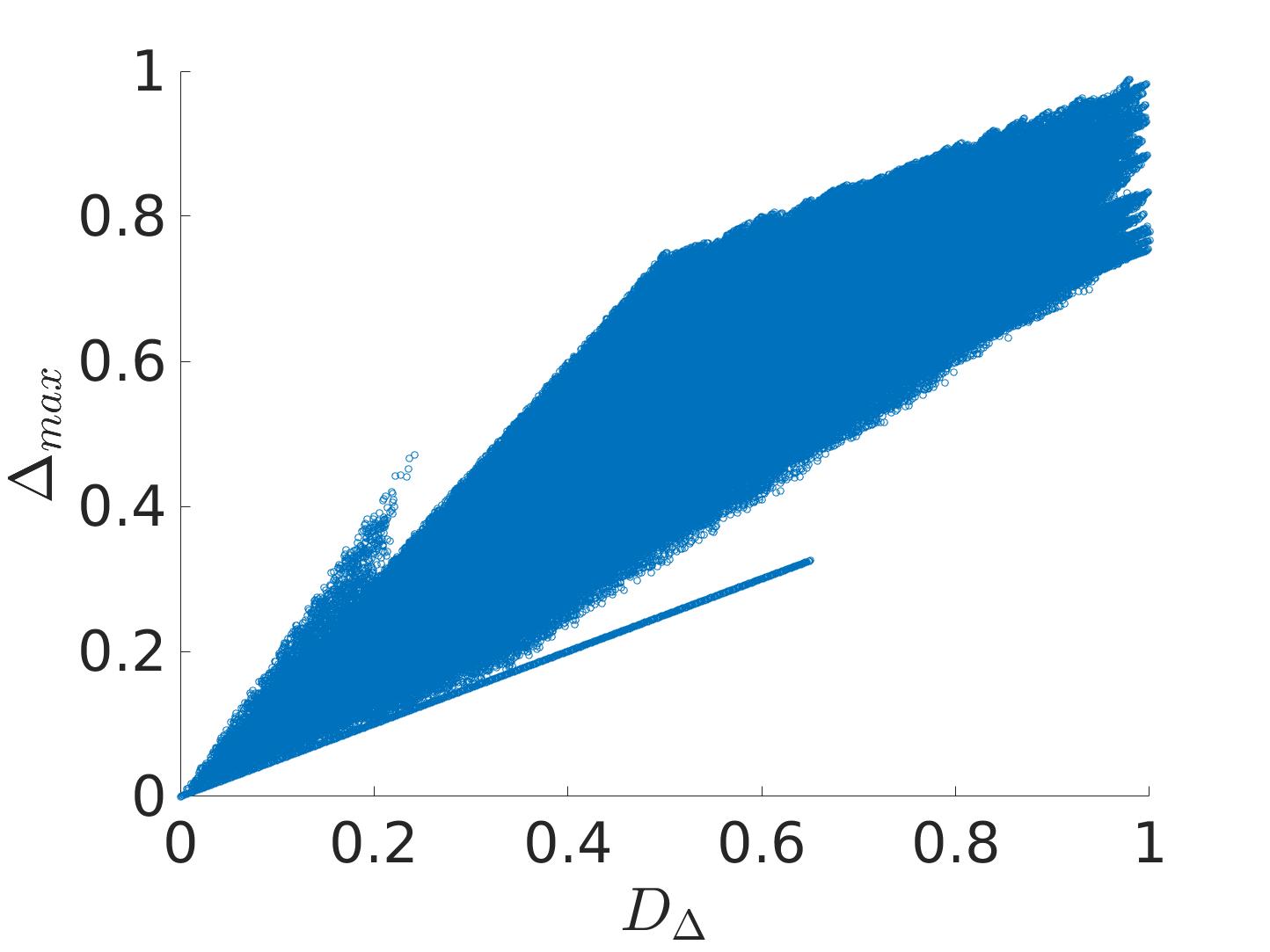}
\caption{Scatter plot of $\Delta_{max}$ against Delta divergence $D_{\Delta}$, comparing the outputs of two classifiers sampled from a population of probability distributions defined over three classes.   } \label{heuristic12}
\end{figure}

In Figures \ref{heuristic21}, \ref{heuristic22}, \ref{heuristic11}, \ref{heuristic12} we present the scatter plots showing the relationship of  $\Delta^*$ and $\Delta_{max}$ to $D_{\Delta}$ for the cases of the three class and six class problems. The first two figures show that $\Delta^*$ is never greater than $D_{\Delta}$. This means that its clutter properties are even better than those of $D_{\Delta}$. Thus the advantage of $D_{\Delta}$ being a proper divergence has been achieved at the expense of slight deterioration of its sensitivity to clutter. 

In the case of the scatter plots for $\Delta_{max}$, it is notable that the variability of this measure is much greater than that of $\Delta^*$. In particular,  for a given $D_{\Delta}$, the values of $\Delta_{max}$ can be both greater or lower than $D_{\Delta}$. Thus $\Delta_{max}$ is less crisp, i.e. in its measurement space there is a greater ambiguity (overlap) between the states of classifier congruence and incongruence. In other words there will be a greater propensity for incongruence detection error. Comparing the six class plot with the three class one suggests that this ambiguity grows with the number of classes.

\section{Conclusions}
\label{conclusions}

The problem of detecting classifier incongruence was addressed in the paper. It involves comparing the output of two classifiers to gauge the level of agreement in their support for a particular decision. As, in general, the output of a classifier is a probability distribution over the admissible hypotheses, classifier incongruence detection basically involves a comparison of these distributions.    
The existing classifier incongruence measures advocated in the literature include the Bayesian  surprise (K-L divergence) \cite{Itti-cvpr05}, or the Delta measures ($\Delta^*$ and $\Delta_{max})$ introduced in \cite{Kittler-pami2013,Kittler-2015}. Unfortunately, the former has a number of undesirable properties and the latter two are heuristic.

Measuring differences between two probability distributions is a standard problem in information theory and statistics. The key tool for this purpose is divergence. Many different divergence functions have been proposed in the literature, each exhibiting different properties. In order to adopt or develop a suitable measure for detecting classifier incongruence it is of paramount importance to understand the properties required for this particular application. We argued that a classifier incongruence measure should focus on differences in the classifier support for the dominant hypotheses, be bounded, symmetric, insensitive to surprisal, and insensitive to clutter induced by nondominant hypotheses. 

The list of required properties postulated in the paper can be considered as  an important contribution in its own right. However, in the context of the paper, this was just a prerequisite for the main task of developing a principled method of measuring classifier incongruence.  A review of existing divergences established that none of them fully satisfied the list of requirements. We adopted the total variation divergence as a starting point, because of its insensitivity to surprisal values. We then reformulated the problem of comparing two probability distributions by grouping all the nondominant classes into a single event. This allowed us to develop the total variation measure into a novel divergence, called {\it Delta divergence}, which is classifier decision cognizant. As a result of this reformulation, the proposed measure is less sensitive to clutter induced by nondominant hypotheses. By studying the characteristics of the proposed measure we demonstrated that it satisfied all the required properties. 

Finally, we conducted a number of experiments showing the relationship of the proposed Delta divergence to baseline classifier incongruence measures, and its robustness to clutter. The experiments confirmed its superiority as a measure of classifier incongruence.




\bibliographystyle{plain}
\bibliography{anomaly,defs,refs,survey}
%

%
%
\end{document}